\documentclass[sigconf, authversion, nonacm]{acmart}
\usepackage{algorithm}
\usepackage{algorithm}
\usepackage[noend]{algpseudocode}
\usepackage{amsmath,amsthm}
\usepackage{booktabs}
\usepackage{enumitem}
\AtBeginDocument{%
  }
\setcopyright{none}
\renewcommand\footnotetextcopyrightpermission[1]{}
\begin{document}

\title{Hybrid Approach to Parallel Stochastic Gradient Descent}

\author{Aakash Sudhirbhai Vora}
\affiliation{%
  \institution{School of Computing and Augmented Intelligence \\ Arizona State University}
  \city{Tempe}
  \state{AZ}
  \country{USA}
}
\email{avora8@asu.edu}

\author{Dhrumil Chetankumar Joshi}
\affiliation{%
  \institution{School of Computing and Augmented Intelligence \\ Arizona State University}
  \city{Tempe}
  \state{AZ}
  \country{USA}
}
\email{djoshi12@asu.edu}

\author{Aksh Kantibhai Patel}
\affiliation{%
  \institution{School of Computing and Augmented Intelligence \\ Arizona State University}
  \city{Tempe}
  \state{AZ}
  \country{USA}
}
\email{apate160@asu.edu}


\begin{abstract}
Stochastic Gradient Descent is used for large datasets to train models to reduce the training time. On top of that data parallelism is widely used as a method to efficiently train neural networks using multiple worker nodes in parallel. Synchronous and asynchronous approach to data parallelism is used by most systems to train the model in parallel. However, both of them have their drawbacks. We propose a third approach to data parallelism which is a hybrid between synchronous and asynchronous approaches, using both approaches to train the neural network. When the threshold function is selected appropriately to gradually shift all parameter aggregation from asynchronous to synchronous, we show that in a given time period our hybrid approach outperforms both asynchronous and synchronous approaches. 
\end{abstract}

\keywords{Data parallelism, Synchronous approach, Asynchronous approach, Stochastic gradient descent, Distributed optimization}
\maketitle
\pagestyle{plain} 

\section{Introduction}
Neural networks play an important role in modern computer applications. They are indispensable from day to day life today. They are the backbone of recommender systems, stock trading algorithms, voice assistants, autonomous driving, etc. Neural networks are trained using Gradient Descent, where the motive is to minimize the loss function by improving model parameters, iterating over the training data. Training the neural network on the whole dataset as input once is a slow process since the training dataset could be huge. An alternative to this is Stochastic Gradient Descent, where data is passed in batches to train the network. So, Stochastic Gradient Descent is widely accepted as the method to train neural network because it is more efficient. 

This process of training neural networks can be accelerated even further by using distributed training where there are multiple worker nodes updating the model parameters in parallel. There are multiple types of parallelism techniques used for training a model. Model parallelism involves splitting the model across different workers, so that each worker control over some parameters of the model. Each worker finds the parameters for the part of model it has and then all the parameters are aggregated to get the final model.Petuum proposed by Xing et .al (2015) \cite{petuum}, Project Adam proposed by Chilimbi et. al (2014) \cite{adam} and Sandblaster proposed by Dean et. al (2012) \cite{dist_belief} all use model parallelism to speed up neural network training.

Another approach that is used for parallelizing training is known as pipeline parallelism. Fundamental idea of pipeline parallelism to overcome the shortcoming of model parallelism that when a  set of workers compute weights for one layer, other workers should not be sitting idly. So, apart from splitting the model, multiple inputs are supplied to the system so that at each point of time, each worker is computing the weights based on one set of inputs. Each worker node is responsible for computing weights for one or more layers. Once the worker computers parameters for the assigned layer, it passes that information to next set of workers responsible for other layers in the network. The same worker is responsible for computing the weights during both feed forward and backpropagation phases of training the model. Model gets divided among workers in the direction of data flow. GPipe proposed by Huang et. al (2019) \cite{gpipe} and Pipedream proposed by Narayanan et. al (2019) \cite{pipe_dream} both make use of pipeline parallelism to speed up the training process.

Data parallelism is another approach that people use for speeding up neural network training. Core idea for this technique is that the training dataset is split among different workers and each worker trains the model in parallel based on the dataset it has. Thereafter, the parameters are combined from the workers to get the resultant model.  Here for parameters to converge, they need to combined. Currently there are two approaches to applying data parallelism, one is synchronous approach. Here, the workers working in parallel need to communicate among themselves to keep the parameters synchronized. One way of synchronizing parameters is to keep the parameters synchronized after each iteration. Another approach is to synchronize parameters after some fixed number of iterations. Popular implementations of synchronous approach to data parallelism are Petuum proposed by Xing et .al (2015) \cite{petuum}, Horovod proposed by Sergeev et .al (2018) \cite{horovod} and Stale Synchronous Parallel Parameter Server proposed by Ho et .al (2013) \cite{ssp}. There is a clear drawback to this approach where faster workers will have to wait for slower workers to catch up. This leads to wastage of resources since faster workers will be sitting idle while slower workers catch up.There is also communication overhead in this approach as workers will coordinate amongst themselves to keep the parameters synchronized. Alternative to this, is the asynchronous approach. Here, workers do not have to communicate amongst each other to keep parameters synchronized. Each worker is allowed to update the parameters independently with little communication needed. This will lead to better resource utilization since the faster workers will not have to wait for slower workers. Popular implementations of this are Hogwild proposed by Niu et .al (2011) \cite{hogwild}, Project Adam proposed by Chilimbi et .al (2014) \cite{adam} and Distbelief proposed by Dean et .al (2012) \cite{dist_belief}. However, there is a major drawback regarding parameter convergence. If the model parameters are not sparse, convergence might take a lot of time since faster workers would frequently update the parameters while slower workers would still be operating on stale parameters. 

We propose a hybrid approach which incorporates advantages of both approaches. This involves using both synchronous and asynchronous approaches. Initially, each worker is computing parameters asynchronously. However as iterations progress, more and more parameters are allowed to be accumulated and passed onto the workers. This approach allows workers to use an asynchronous approach to get more progress per iteration and a synchronous approach to get more confident progress. 

\section{Related Work}
To increase the effectiveness of neural network training over large training sets, a number of strategies have been put forward. Due to its effectiveness in handling huge datasets, stochastic gradient descent (SGD) has been largely favored over gradient descent. In order to increase efficiency even more, distributed training techniques have been investigated. These techniques let worker nodes update model parameters concurrently.

Synchronous parallel gradient descent is one of the current training methods for neural networks. This method synchronizes all worker nodes either instantly or after a predetermined number of iterations. The output parameters are closely matched to those acquired by sequential stochastic gradient descent thanks to this synchronization. This strategy has a big drawback in that, faster workers must wait for slower workers to catch up, which adds a lot of idle time. The work "More Effective Distributed ML via a Stale Synchronous Parallel Parameter Server" \cite{ssp} offers a synchronous parallel gradient descent approach with a parameter server, with updates that are slightly stale. However, the disadvantage is that faster workers need to wait for slower workers to catch up, resulting in idle time. The publication "Horovod: Fast and Easy Distributed Deep Learning in TensorFlow" \cite{horovod} describes a distributed training system that employs synchronous parallel gradient descent in conjunction with a ring-based communication technique called ring-all reduce. With features like gradient compression, it enhances scalability and speed. Fast and efficient training is one of the benefits, although it shares the disadvantage of idle time with slower workers. The paper "Petuum: A New Platform for Distributed Machine Learning on Big Data" \cite{petuum} introduces Petuum, a distributed machine learning platform that uses synchronous parallel gradient descent with a parameter server. It makes use of techniques like model compression and a customized communication protocol. Large-scale machine learning jobs can be handled efficiently, but it also suffers from idle time because of slower workers.

Asynchronous parallel gradient descent is another contemporary strategy in which each worker separately updates parameters in parallel. Because workers do not have to wait for each other, this strategy optimizes the utilization of resources. However, convergence will only be achievable for sparse modifications, and individual worker parameters could become outdated.  The paper, "HOGWILD!: A Lock-Free Approach to Parallelizing Stochastic Gradient Descent," \cite{hogwild} introduces a technique that uses a lock-free mechanism to parallelize stochastic gradient descent. It enables independent model parameter updates across several threads without explicit synchronization.  While this approach effectively parallelized SGD minimizing synchronization overhead and resource usage, the method's assumption of sparse updates makes it potentially inconclusive for dense updates. Additionally, because there is no explicit synchronization, individual threads may use outdated parameters, which could have an effect on the training process convergence and accuracy.  "Project Adam: Building a Scalable and Efficient Deep Learning Training System" \cite{adam} also uses asynchronous parallel SGD to allow worker nodes to update simultaneously. To share information among workers and maintain uniformity, it makes use of a communication protocol. Individual workers' use of out-of-date criteria can have an impact on how well training is performed generally, sometimes producing less-than-ideal results. Existing techniques exhibit various trade-offs in terms of synchronization and convergence assurances. 

\section{Problem Formulation}
Let there exist $X$ which is a set of points sample over some unknown distribution defined over the space of $n$ dimension i.e. $R^n$. Corresponding to each point $x \in X$, there exists a value $y$ sampled from another set $Y$ which is a subset of $R^m$. The goal is to find a function $f:R^n \times R^k \rightarrow R^m$ defined on $X$ and parameter $\theta$, where $\theta \in R^k$ such that it can approximate the unknown function $f_{un}:R^n \rightarrow R^m$ which maps each point in $x$ to corresponding $y$. Further, we assume that there exists a differentiable convex function $j(x,y,\theta)$ defined over $R^n \times R^m \times R^k \rightarrow R$. Let a new function $J$ be defined for any subset of $X$ which applies $j$ to all $x \in X$ and returns a summation of them. The function $f$ can be estimated by finding the value of the parameter theta which minimizes the function $J(X,Y,\theta)$. This minimization can be performed by gradient-based methods like Stochastic Gradient Descent.

\begin{equation}
  J(X,Y,\theta) = \Sigma_{x_i \in X, y_i \in Y} j(x_i,y_i,\theta)
\label{eq1}
\end{equation}
\begin{equation}
  \theta = argmin_{\theta \in R^k} J(X,Y,\theta)
\label{eq2}
\end{equation}
\begin{equation}
  \theta_i' = \theta_i - \eta \frac{\partial J(X,Y,\theta)}{\partial \theta_i}
\label{eq2}
\end{equation}

However, we have multiple processors at our disposal that we would like to use for performing the minimization. We assign one of the machines to be in charge of maintaining the current values of the parameter and call it the Parameter server p. Rest of the machines are known as worker machines $w_i$. Each of the worker machines $w_i$ has a subset of data $(X_i,Y_i)$ from the entire dataset $(X,Y)$. They are responsible for getting the current parameter values, calculating the gradient using $(X_i,Y_i)$, and sending them to the Parameter server $p$. We don't assume any kind of global notion of time among these workers. Furthermore, we assume that the execution speed of each worker is different and there also exists communication delays when workers are communicating with the Parameter Server. 

In such a setting, it becomes necessary to define whether the policy is used by the Parameter server for updating the parameters from the gradients received from the worker. Two widely used approaches are synchronous and asynchronous. In the synchronous approach, the parameter collects gradients from all workers before updating the parameter and each worker waits till updated parameters are not received. Each update in this approach is calculated by all workers on the same previous values of parameters and hence the updates are less noisy. However, this algorithm is slower since faster workers have to wait for slower ones.  On the other hand, in an asynchronous approach, any parameter applies the update as soon as it is received from the worker and provides a new value of the parameter to the worker. Though this approach is simple and works very well in practice during the initial convergence(reach quickly to local minima), as we move closer to local minima, faster workers might end up providing too many stale updates which might lead to noisy updates and slow down convergence. Thus, we ask the question, is there a way that combines good properties from both synchronous and asynchronous algorithms i.e. high confidence updates and faster initial convergence respectively, and gives a new algorithm? We hypothesize that if we start initially with asynchronous updates and switch over time to synchronous updates, it would lead to faster convergence.

\section{Methodology: Smooth Switch Algorithm}

To develop a hybrid algorithm, which we call a smooth switch algorithm, for stochastic gradient descent (SGD) algorithms, both synchronous and asynchronous techniques are combined. With this method, asynchronous updates are used to increase iteration throughput while synchronous updates are used to ensure the integrity of forward positive progress. The method defines a smooth transition from the asynchronous to the synchronous approach which is controlled by the threshold function. Asynchronicity will help in achieving more updates per iteration and synchronicity will help improve accuracy per iteration. This two-pronged method offers a fair trade-off between training speed and accuracy by combining the advantages of synchronous and asynchronous updates.
\begin{figure}[h!]
    \centering
    \includegraphics[width=8.5cm,height=6.5cm]{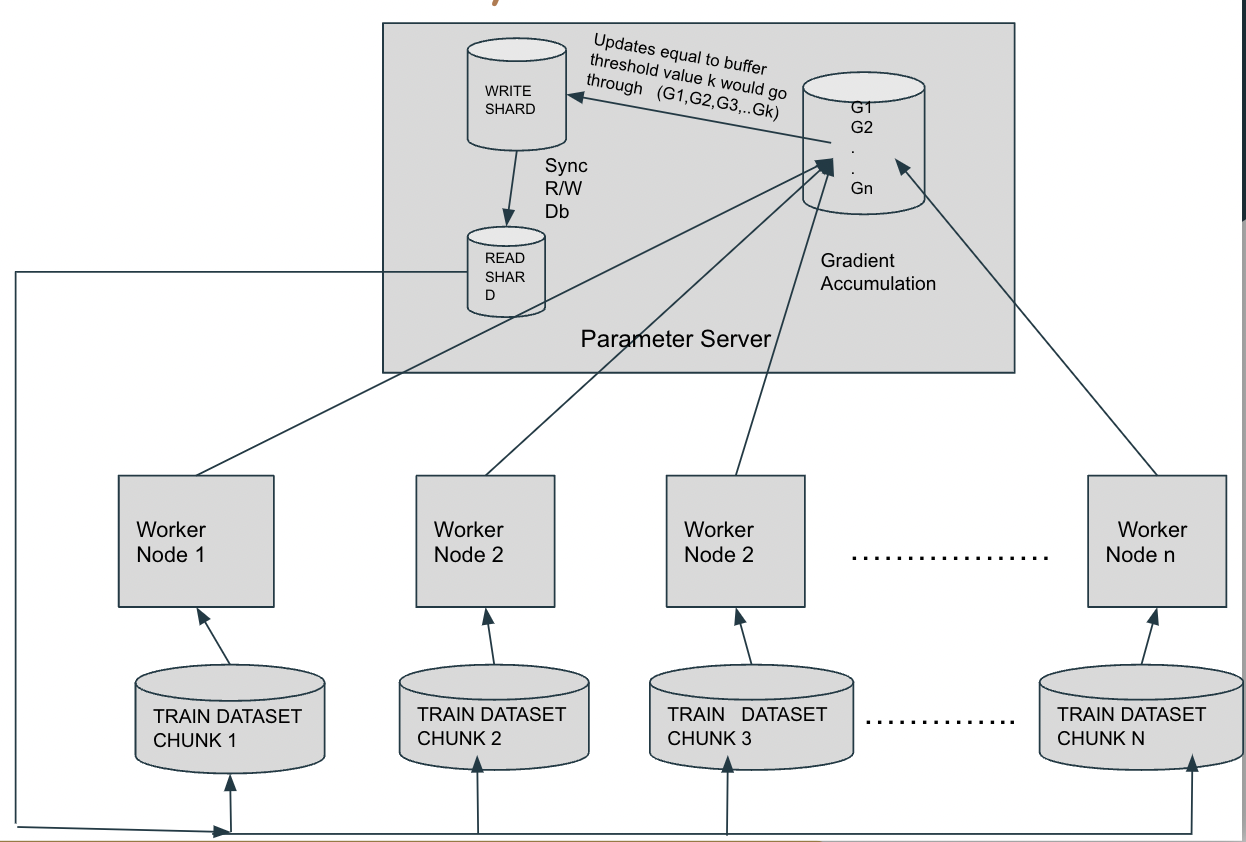}
    \caption{System Architecture}
    \label{fig:architecture}
\end{figure}
Figure \ref{fig:architecture} displays the System Architectural diagram in which a threshold parameter (K) is used to control the transition from asynchronous to synchronous updates. Initially, the threshold is set to a low value, enabling asynchronous updates to the parameter server. As the training process iterates, the threshold gradually increases. During each iteration, multiple gradients (G1, G2, G3, ... Gk) are accumulated in the gradient buffer, and at each iteration, the algorithm evaluates if the gradients accumulated are greater than or equal to the threshold to trigger the transition to synchronous updates to the parameter server. If the threshold has not been reached, asynchronous updates continue. Once the threshold is reached, synchronous gradient updates are performed to the parameter server. The steps are repeated until convergence or a specified number of iterations. The algorithm \ref{alg:algorithm} demonstrates the entire process.
\begin{algorithm}[h!]
    \caption{}
    \label{alg:algorithm}
    \begin{algorithmic}
        
        \State Step 1: Set the initial gradient buffer to the parameter database transfer threshold (K) to a very low value to allow for asynchronous updates (stale reads).   
        \State Step 2: \While {(!(convergence) or !(a specified number of iterations))}
        \EndWhile
         \begin{enumerate}
        \item  If the total gradients in the gradient buffer >= threshold K then synchronize all the gradients in the gradient buffer with the Parameter Server (reduction in stale reads).
        \item If the threshold has not been reached, continue with asynchronous updates to the Parameter Server. (Might cause frequent stale reads)
        \item Gradually increase the threshold (K) as the iterations progress.
        \item Accumulate multiple gradients (G1, G2, G3, ... Gn) during each iteration.
    \end{enumerate}

    \end{algorithmic}
\end{algorithm}

\section{Data Preparation}
\begin{figure}[!h]
    \centering
    \includegraphics[width=8cm,height=3.5cm]{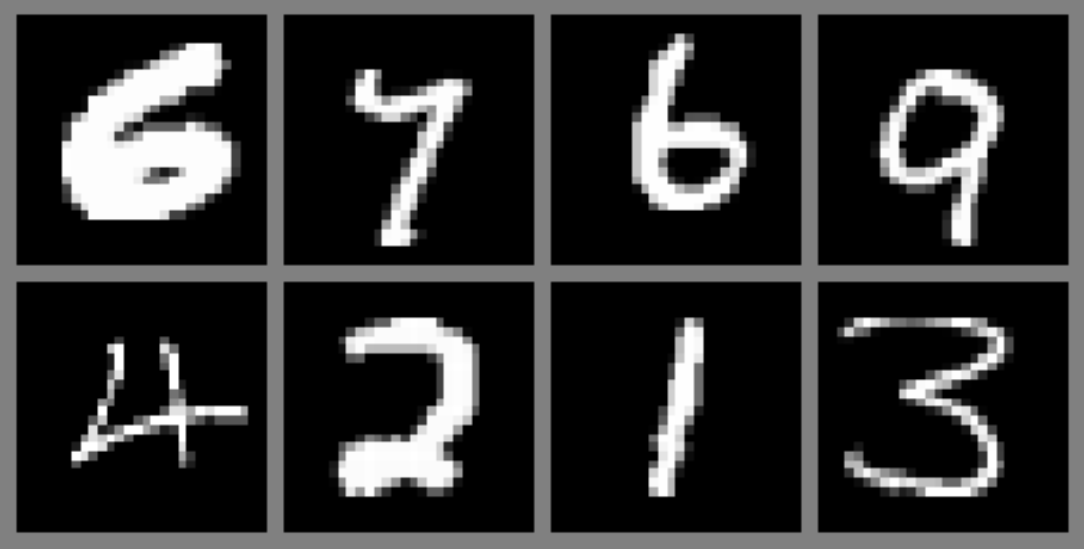}
    \caption{MNIST samples}
    \label{fig:mnist handwritten}
\end{figure}

\begin{figure}[!h]
    \centering
    \includegraphics[width=8cm,height=6.5cm]{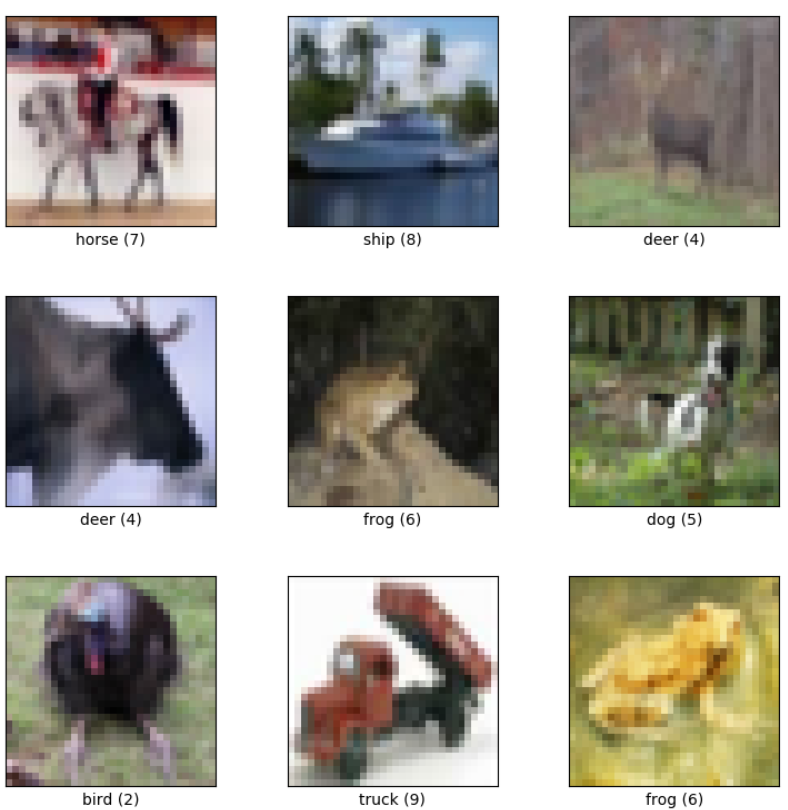}
    \caption{CIFAR 10 samples}
    \label{fig:cifar10image}
\end{figure}

As part of our research experiment, we conducted evaluations to assess the performance and generalization capabilities of our approach to Stochastic Gradient Descent (SGD) using a diverse range of datasets. Among these datasets, we utilized the MNIST dataset \cite{yannlecunn} figure \ref{fig:mnist handwritten} and the CIFAR-10 dataset \cite{cifarcitation} figure \ref{fig:cifar10image}.

A curated collection of grayscale images showing handwritten numbers makes up the MNIST dataset. It serves as a well-known benchmark dataset for tasks involving image classification in which the goal is to correctly identify the represented digit in each image. The MNIST dataset is a rich source of labeled data with 60,000 training images and 10,000 test images, providing a significant quantity of resources for testing and training machine learning algorithms. With a size of 28x28 pixels for each image, the collection has 784 features. The MNIST dataset has been widely used to evaluate the effectiveness of various machine learning models due to its well-stated task and simplicity.
In addition, the CIFAR-10 dataset \cite{cifarcitation} included in our research comprises a total of 60,000 color images that make up the CIFAR-10 dataset and are split into 10,000 test images and 50,000 training images. These photos, which represent numerous objects and scenes, are divided into ten different classifications. The CIFAR-10 dataset is more sophisticated than MNIST since it has a resolution of 32x32 pixels and three RGB color streams. It is regularly used as a benchmark for assessing how well image classification systems can handle increasingly challenging visual tasks.

We wanted to evaluate the performance and generalization capabilities of our model using both the MNIST and CIFAR-10 datasets. We were able to assess the performance of our method using a variety of datasets, including basic grayscale photographs of numbers (MNIST) and more complex color photos of a variety of objects (CIFAR-10). Through this investigation, we gained valuable insights into the adaptability and robustness of our distributed large-scale Stochastic Gradient Descent approach.

\section{IMPLEMENTATION AND EXPERIMENTS}

The goal of our experiment is to validate our hypothesis that the proposed algorithm provides a speedup in convergence as compared to both completely synchronous and asynchronous versions of distributed stochastic gradient descent. Hence, in all our experiments, we ran all three algorithms namely, our proposed algorithm, synchronous and asynchronous algorithms for the same initial conditions, and collected values of training loss, testing loss, and testing accuracy at various time intervals.

All our experiments were executed in a clustered environment setup. The cluster was provided with 2 CPUs each with 14 cores which makes 28 cores in total. The CPUs were based on 64-bit x86 architecture and the model name was Intel(R) Xeon(R) CPU E5-2680 v4 with 2.4GHz clock speed. The kernel version was 3.10.0-1160.21.1.el7.x86\_64 and the operating system was CentOS Linux 7. The available RAM on the cluster was 263.85 GB.

The code for all algorithms was written in Python language with version 3.9.12. For parallel execution and scheduling, we used the ray library with version 2.4.0. Pytorch 2.0.0 was used for model creation and training. For the training of the model, 25 gradient workers were used for calculating the gradient and they passed the updates to a worker which acted as Parameter Server. Further, during the training, to simulate the communication delays and faster/slower workers, we randomly introduced execution delays in 50\% gradient workers. The execution delays were sampled randomly from a normal distribution with a mean of 0 and a standard deviation of 0.25 during each gradient calculated by the worker.

For testing our approach, we selected MNIST and CIFAR-10 datasets. For our algorithm, we used the step function as our threshold function which updated the threshold with an increase in the number of gradient updates. We executed our experiments for combinations of step sizes in multiples of 3 and 5 of reciprocal of learning rate and batch size of 32 and 64. For each combination, we trained the model for 5 rounds starting from random initialization using our algorithm, asynchronous and synchronous algorithm. For each round, the same initialization values of weights were used for each algorithm. The training was performed for 100 seconds in each round. Since both datasets contain images, CNN was used as the model. For experiments, we fixed the learning rate to 0.01. Since we are creating a model to solve the classification problem, negative log-likelihood loss is used.

Furthermore, we wanted to analyze the effects of various step sizes, batch sizes, and communication delays on our algorithm. Hence, we repeated our experiments for different combinations of step size, batch size, and communication delays and noted training and testing loss along with testing accuracy during different intervals. For this purpose, we used randomly generated datasets with 20 dimensions and 10 classes containing 10k samples with 80:20 train to test split. A newly sampled dataset was used for each configuration. The reason behind selecting a random dataset was to cover a wide range of classification problems and validate the robustness of our algorithm.


\section{RESULTS and DISCUSSION}

In this section, we present the results of various experiments and analyses of the same. First, we present how our algorithm works on the MNIST and CIFAR-10 datasets. After that, we also present the effects of various choices of step sizes, batch sizes, and communication delays on our algorithm.

\subsection{Results on MNIST and CIFAR-10 dataset}
\begin{figure}[!h]
    \centering
    \includegraphics[width=4cm,height=3.5cm]{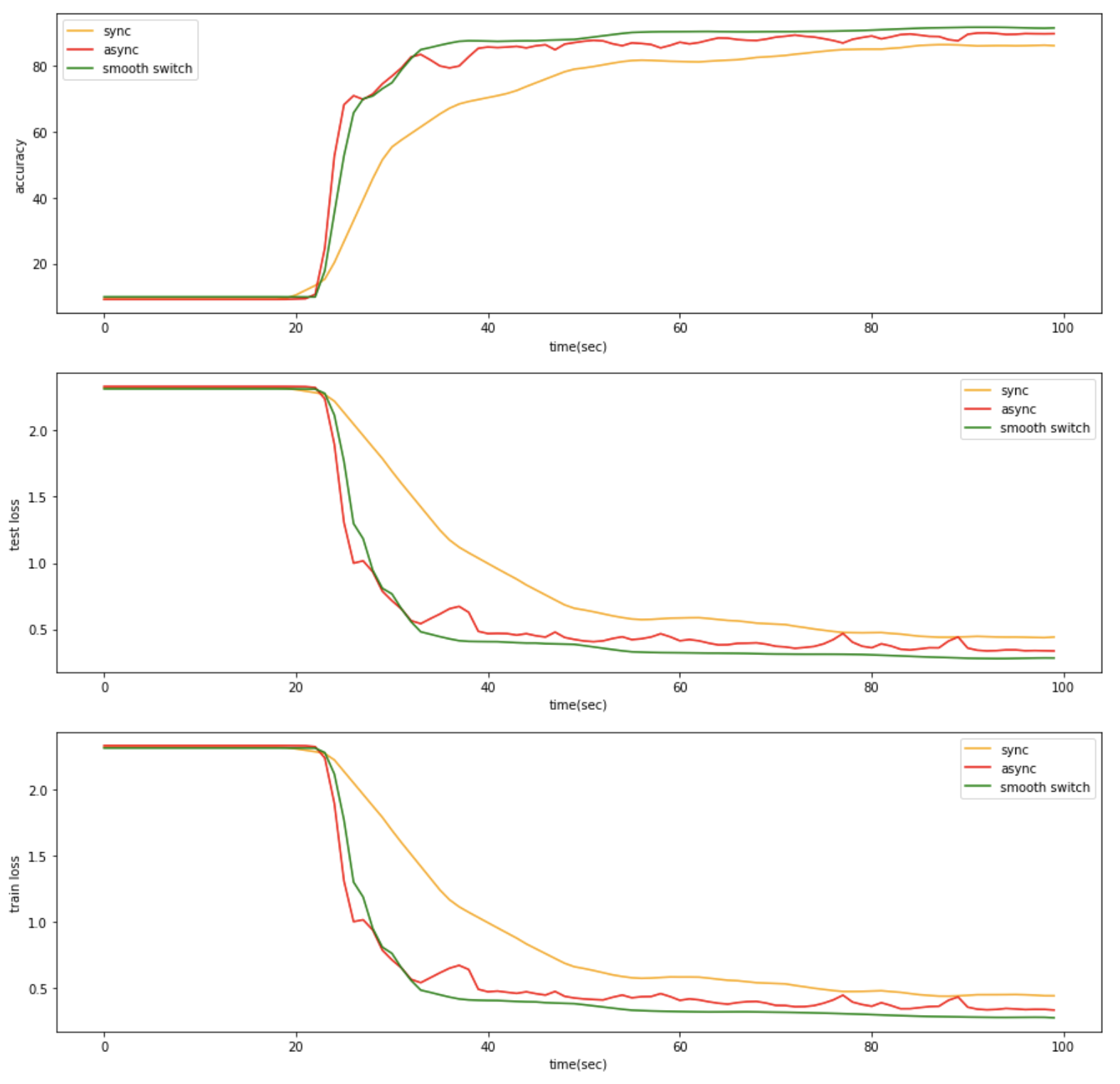}
    \includegraphics[width=4cm,height=3.5cm]{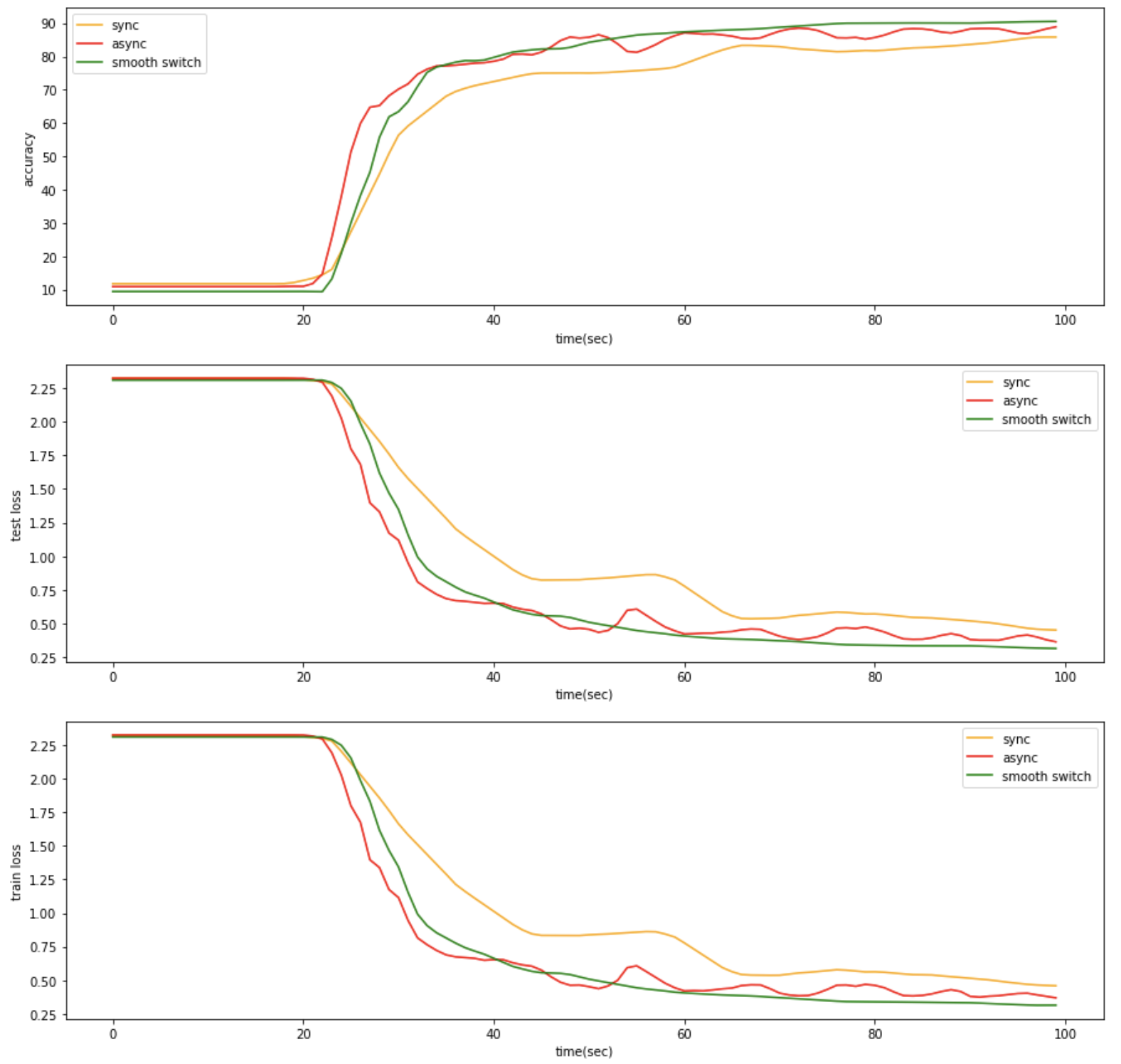}
    
    \caption{Testing accuracy, testing loss and training loss on MNIST for step size 300, and batch size 32(left) and 64(right)}
    \label{fig:mnist3003264}
\end{figure}

\begin{figure}[!h]
    \centering

    \includegraphics[width=4cm,height=3.5cm]{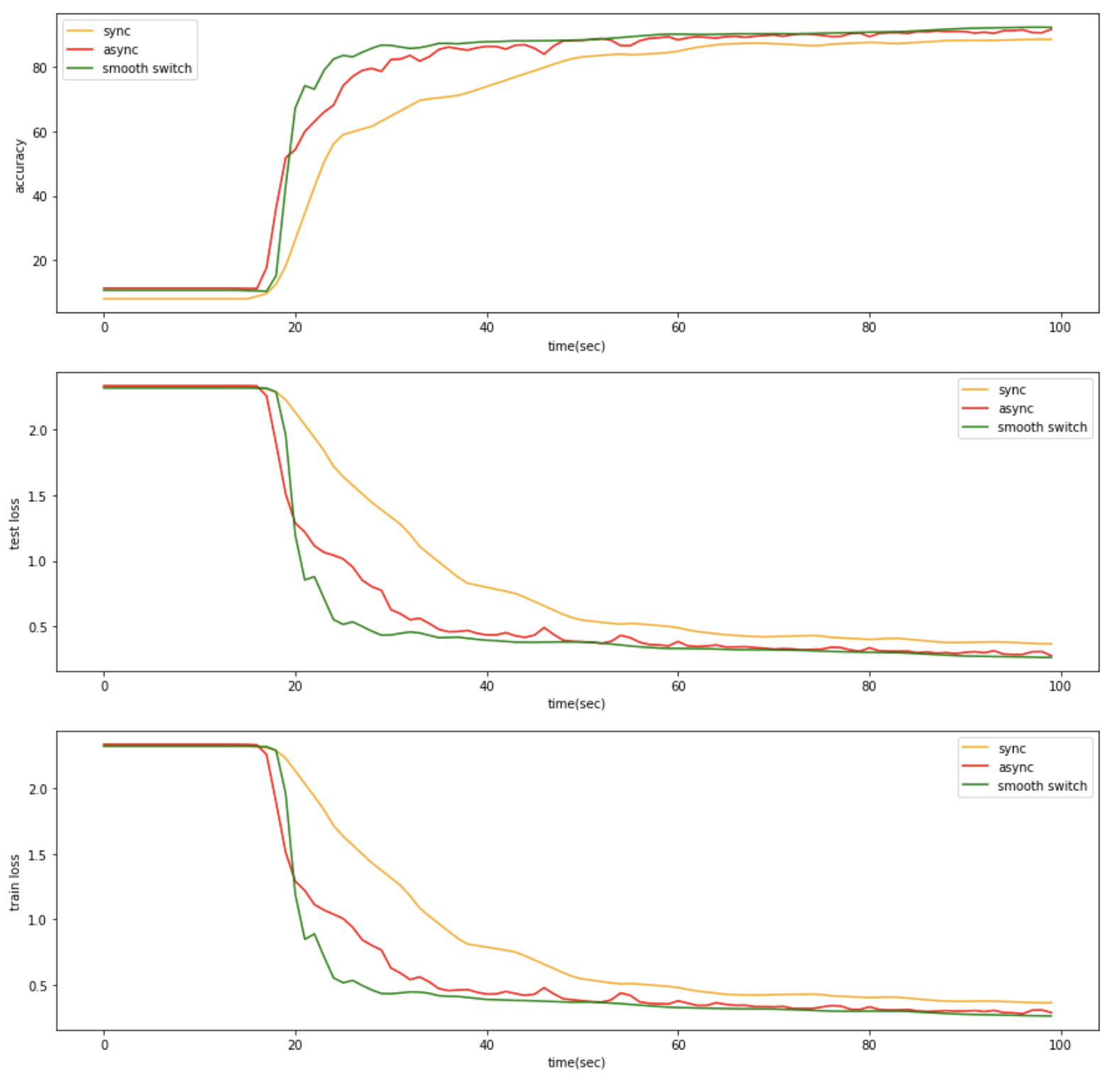}
    \includegraphics[width=4cm,height=3.5cm]{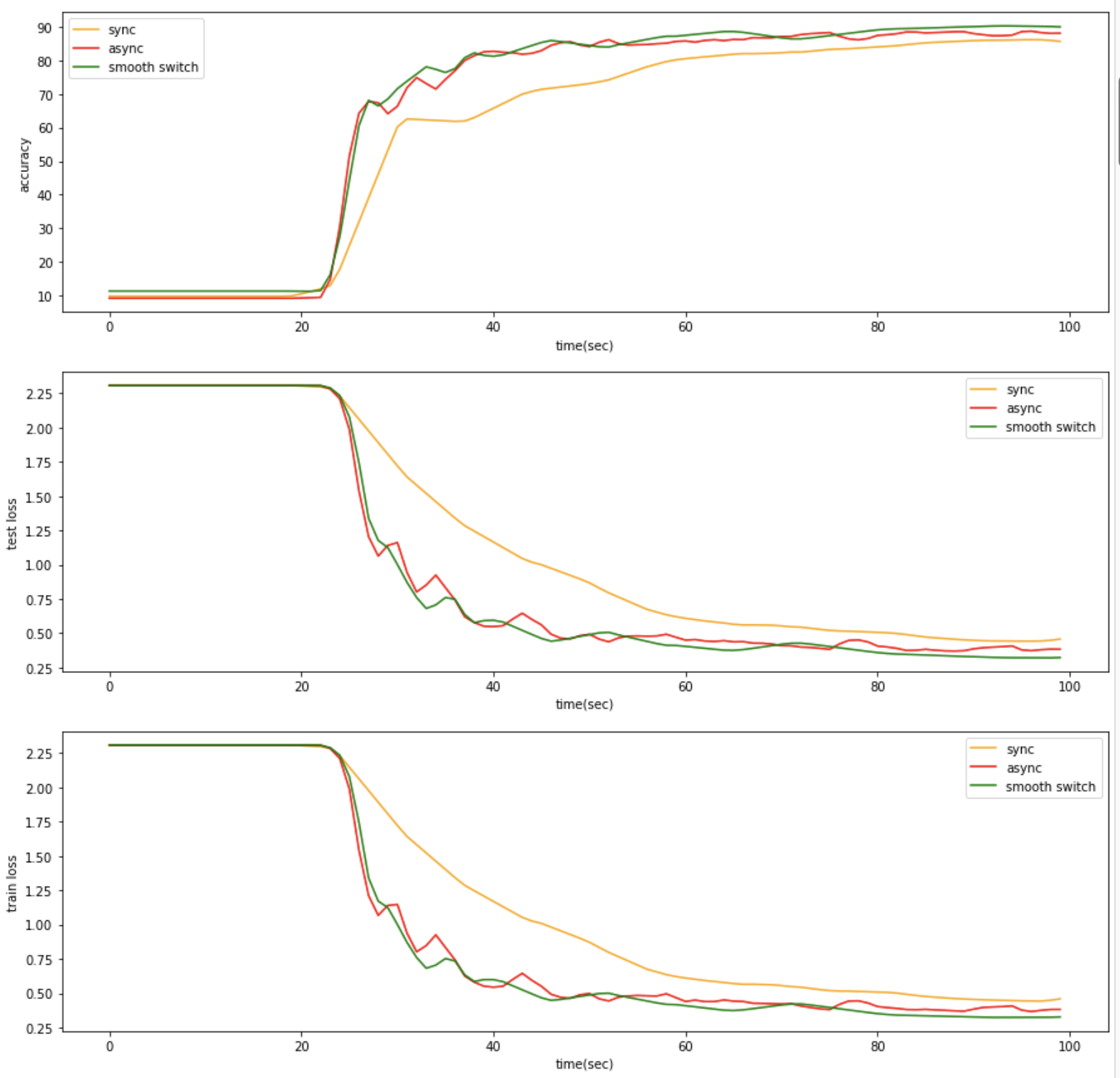}
    
    \caption{Testing accuracy, testing loss and training loss on MNIST for step size 500, and batch size 32(left) and 64(right)}
    \label{fig:mnist5003264}
\end{figure}

Plots \ref{fig:mnist3003264} and \ref{fig:mnist5003264} shows the average values of testing accuracy, testing loss, and training loss for five rounds of training from random initialization on the MNIST dataset. It can be seen clearly that our algorithm maintains the lead in terms of accuracy and loss as compared to both asynchronous and synchronous versions. The same trend is observed for all the combinations of batch sizes and step sizes. However, the speed gain by our algorithm over the asynchronous version is not that significant, we believe that MNIST poses a simple optimization problem that does not bring out problems of asynchronous algorithm effectively. Table \ref{table:mnist} shows the difference of the metrics like accuracy and loss between our algorithm and asynchronous algorithm averaged over the entire training interval. For better performance, the difference in accuracy should be positive and that loss should be negative.

\begin{table}[h!]
\centering
\begin{tabular}{ |c|c|c|c|c| } 
 \hline
 $\frac{(Step size,Batch size)}{Metric}$ & (300,32) & (300,64) & (500,32) & (500,64)\\ 
 \hline
 Test Accuracy & 1.374 & -0.516 & 1.366 & 1.291\\
 \hline
 Test loss & -0.047 & 0.001 & -0.053 & -0.022 \\
 \hline
 Train loss & -0.047 & -0.001 & -0.054 & -0.023 \\
 \hline
\end{tabular}
\caption{Difference between the metric for our algorithm and asynchronous algorithm averaged over entire training interval for MNIST dataset. For better performance, difference in accuracy should be positive and that loss should be negative}
\label{table:mnist}
\end{table}

\begin{table}[h!]
\centering
\begin{tabular}{ |c|c|c|c|c| } 
 \hline
 $\frac{(Step size,Batch size)}{Metric}$ & (300,32) & (300,64) & (500,32) & (500,64)\\ 
 \hline
 Test Accuracy & 4.849 & 2.435 & 3.468 & 2.884\\
 \hline
 Test loss & -0.137 & -0.066 & -0.092 & -0.080 \\
 \hline
 Train loss & -0.139 & -0.067 & -0.091 & -0.082 \\
 \hline
\end{tabular}
\caption{Difference between the metric for our algorithm and asynchronous algorithm averaged over entire training interval for CIFAR-10 dataset. For better performance, difference in accuracy should be positive and that loss should be negative}
\label{table:cifar}
\end{table}

\begin{figure}[!h]
    \centering
    \includegraphics[width=4cm,height=3.5cm]{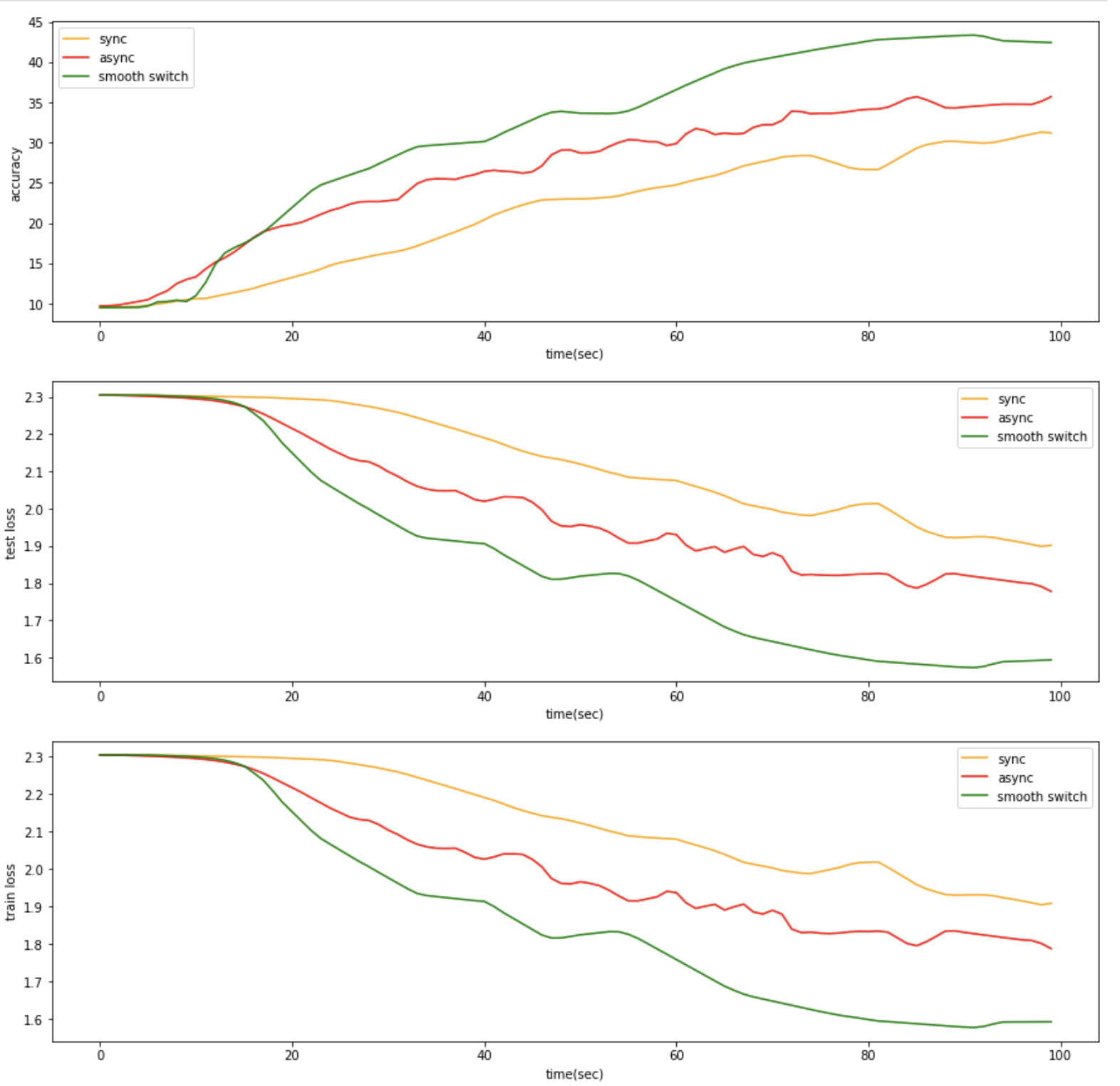}
    \includegraphics[width=4cm,height=3.5cm]{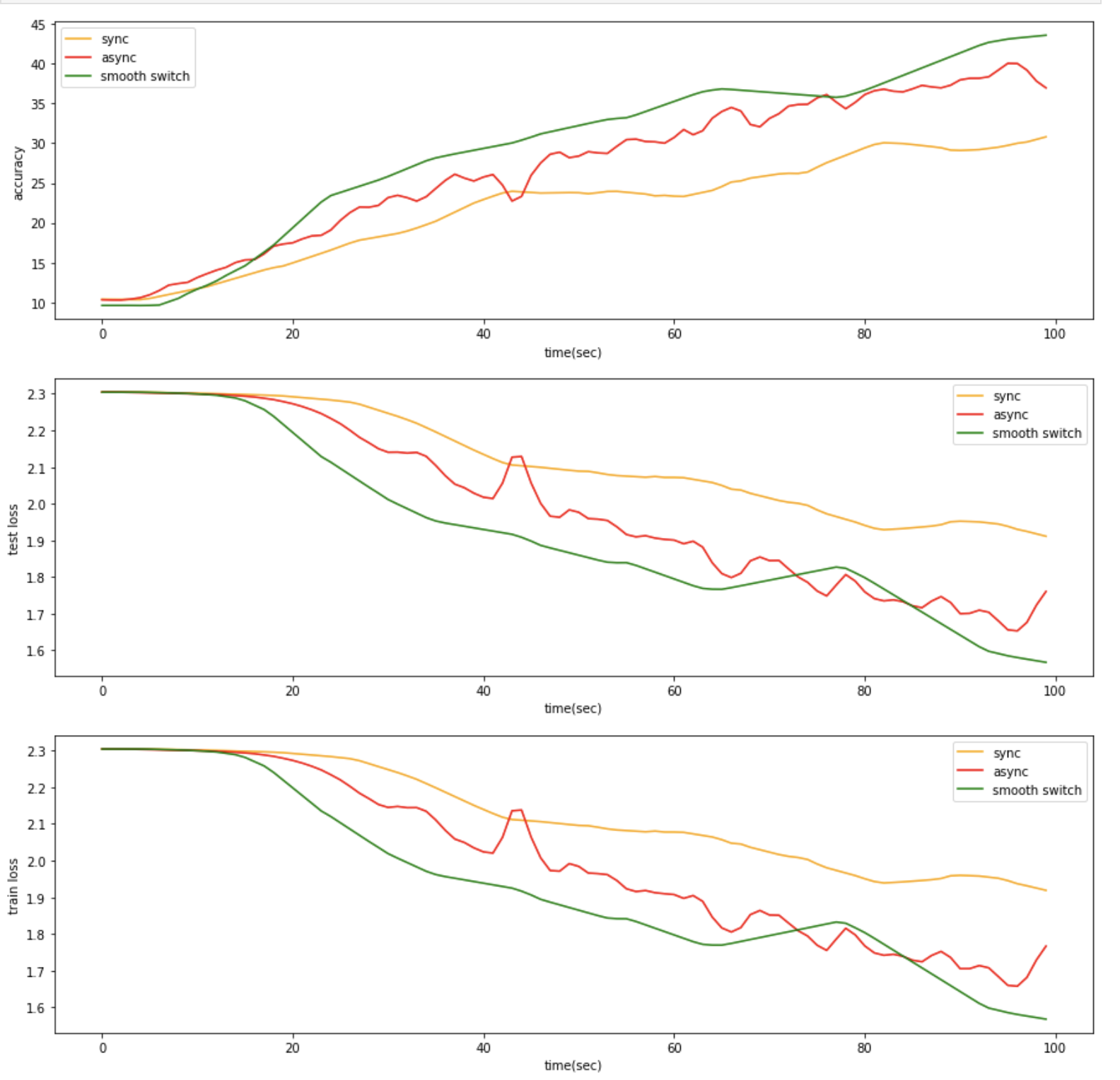}
    
    \caption{Testing accuracy, testing loss and training loss on CIFAR-10 for step size 300, and batch size 32(left) and 64(right)}
    \label{fig:cifar3003264}
\end{figure}

\begin{figure}[!h]
    \centering

    \includegraphics[width=4cm,height=3.5cm]{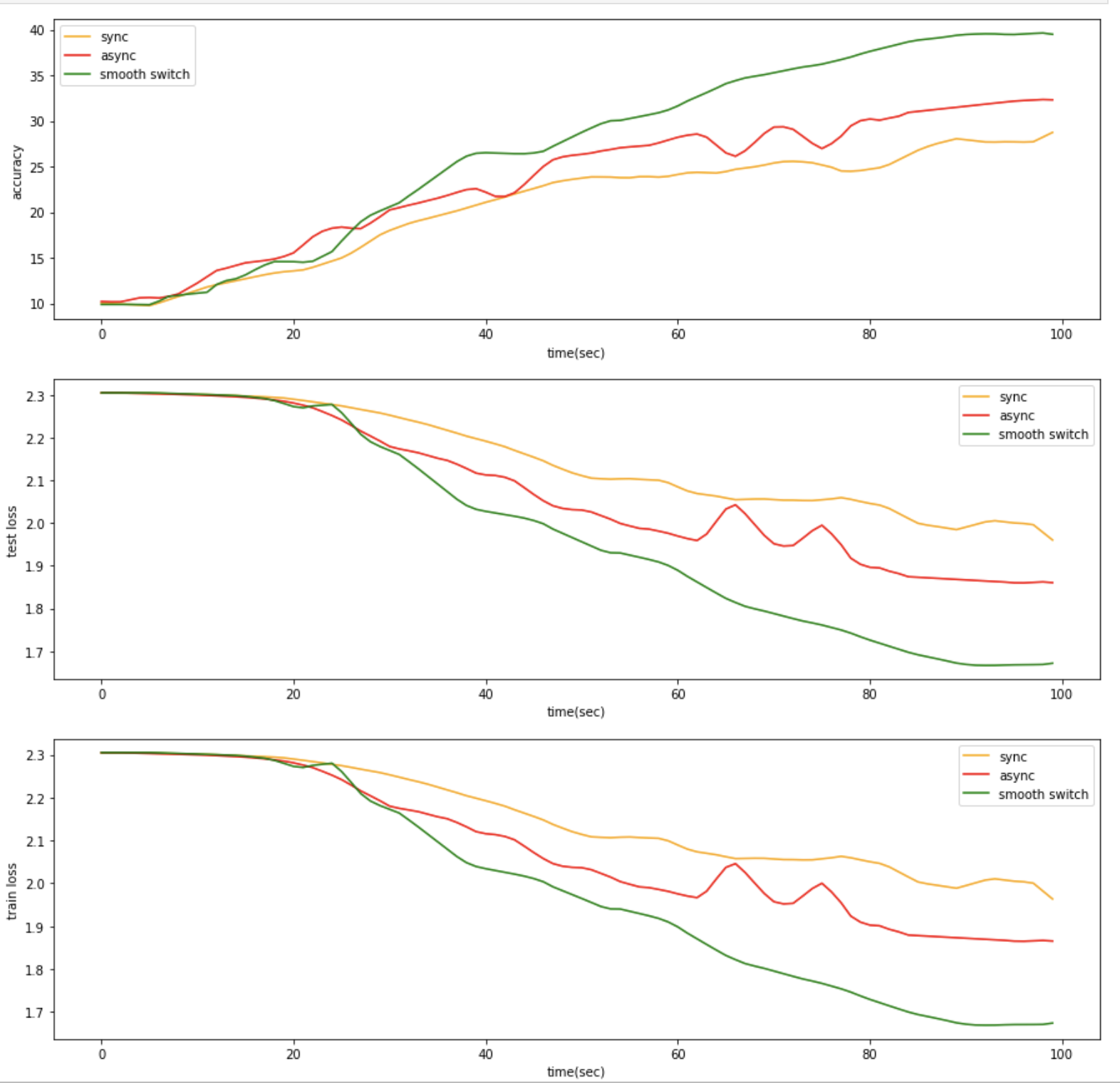}
    \includegraphics[width=4cm,height=3.5cm]{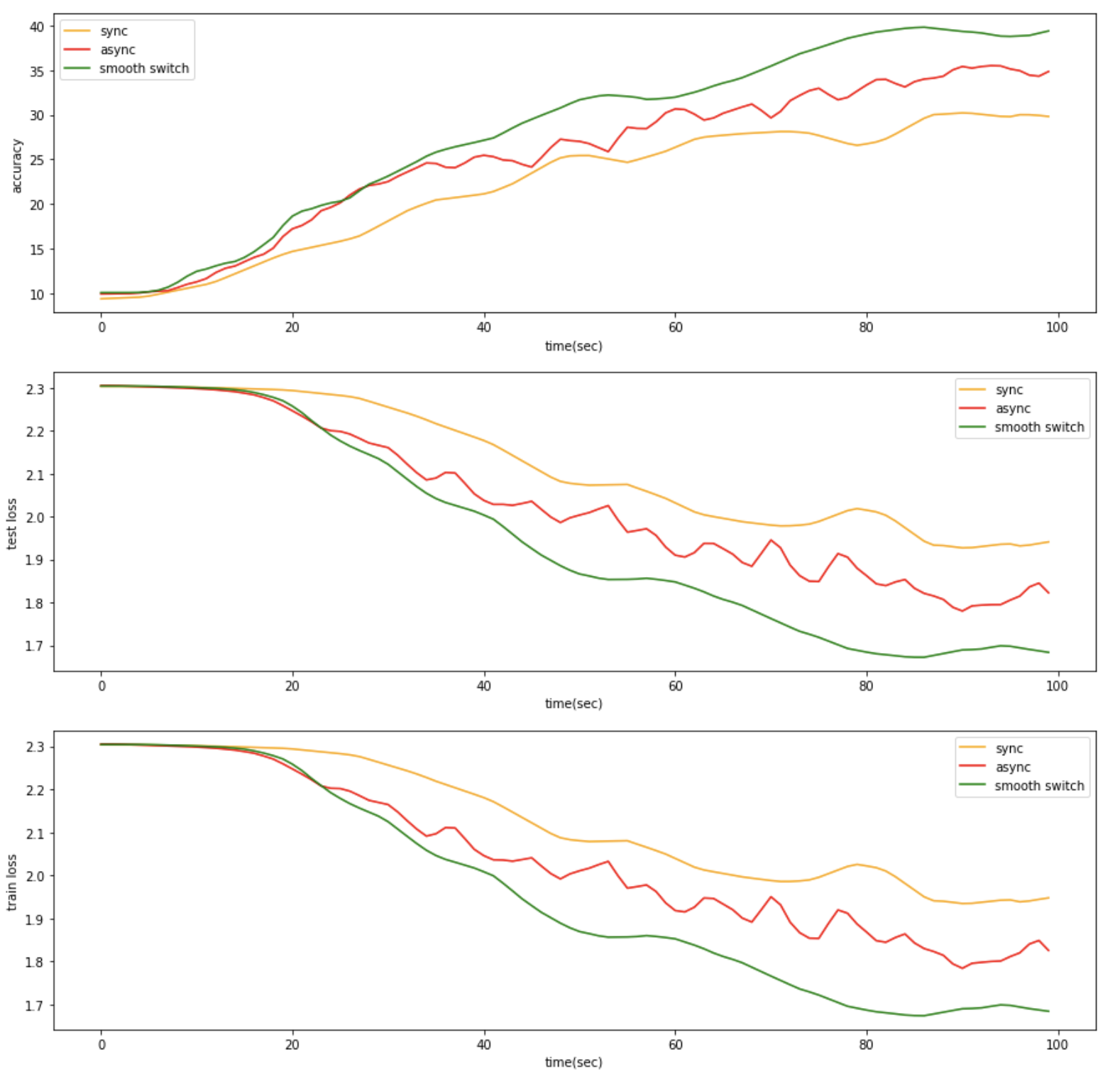}
    
    \caption{Testing accuracy, testing loss and training loss on CIFAR-10 for step size 500, and batch size 32(left) and 64(right)}
    \label{fig:cifar5003264}
\end{figure}

For the next set of experiments, we selected CIFAR-10 as our dataset since we believe that it provides a difficult optimization problem as compared to MNIST. Table \ref{table:cifar} and plots \ref{fig:cifar3003264} and \ref{fig:cifar5003264} show similar statistics as that for MNIST. We can clearly note here that our algorithms show significant speedup as compared to both of the other algorithms. It is able to achieve higher accuracy and lower loss as compared to asynchronous and synchronous algorithms. In all the previous experiments, the synchronous algorithm was very slow, and hence for future analysis, only present a comparison between our algorithm and the asynchronous algorithm.

\subsection{Effect of different batch sizes}
\begin{figure}[!h]
    \centering

\includegraphics[width=8cm,height=6.5cm]{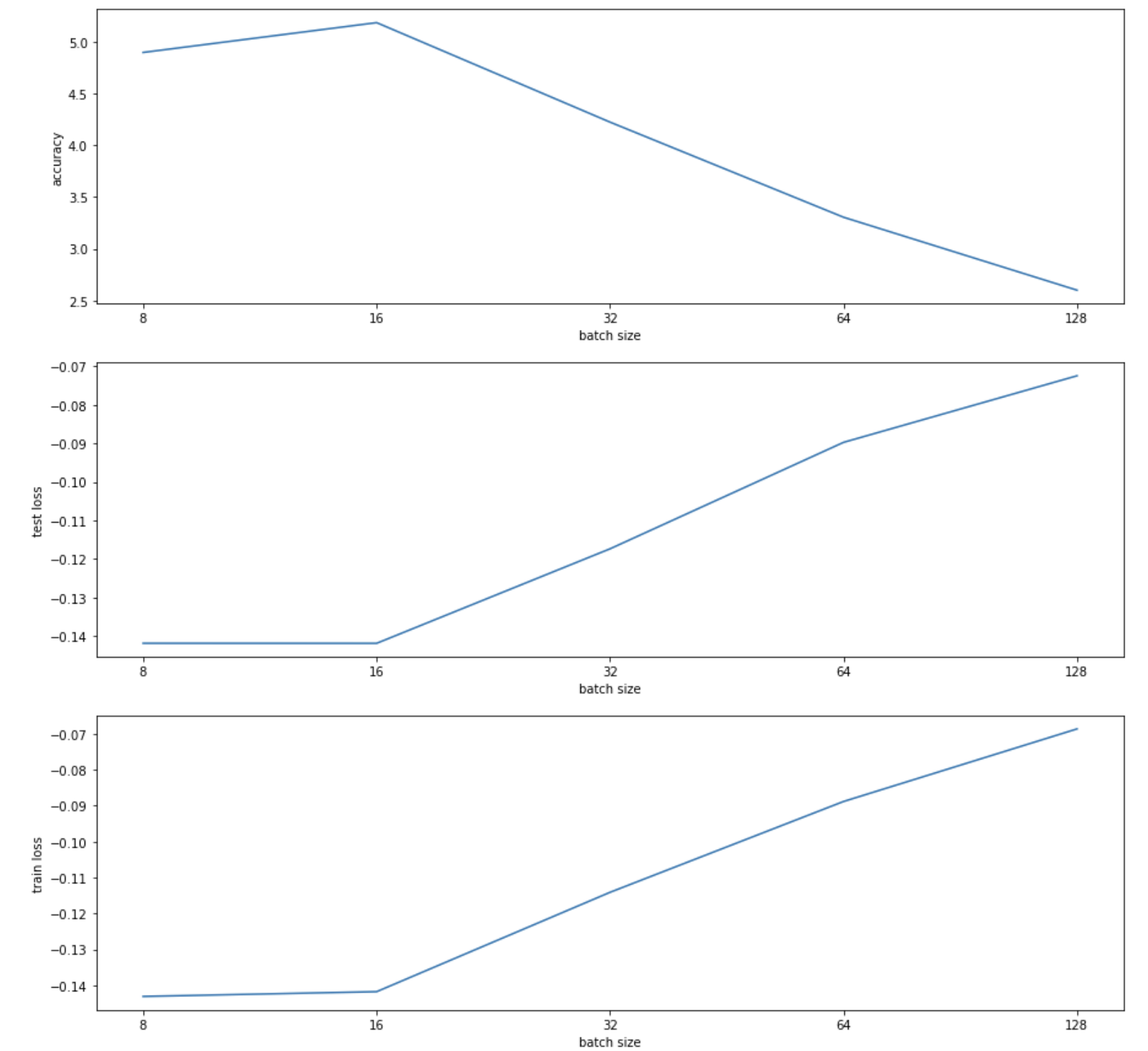}
    
    \caption{Average difference in metrics for different batch sizes}
    \label{fig:batchsize}
\end{figure}

Further, we wanted to understand how different values of batch sizes affect the efficiency of our approach. For each of the batch sizes, we executed 5 rounds of training, each with different initialization of the parameters on the randomly generated dataset. Table \ref{table:batchsize} shows the difference of the metrics like accuracy and loss between our algorithm and asynchronous algorithm averaged over the entire training interval. We hypothesized that as the batch size increases, the difference should decrease since asynchronous algorithms start providing updates with high confidence. This can be also validated by the trend observed in the plot \ref{fig:batchsize}.

\begin{table}[h!]
\centering
\begin{tabular}{ |c|c|c|c|c|c| } 
 \hline
 $\frac{(Batch size)}{Metric}$ & 8 & 16 & 32 & 64 & 128\\ 
 \hline
 Test Accuracy & 4.896 & 5.183 & 4.222 & 3.304 & 2.599\\
 \hline
 Test loss & -0.141 & -0.141 & -0.117 & -0.089 & -0.072 \\
 \hline
 Train loss & -0.143 & -0.141 & -0.114 & -0.088 & -0.068 \\
 \hline
\end{tabular}
\caption{Difference between the metric for our algorithm and asynchronous algorithm averaged over entire training interval for various batch sizes and constant step size of 500. For better performance, difference in accuracy should be positive and that loss should be negative}
\label{table:batchsize}
\end{table}

\subsection{Effect of different step sizes}
\begin{figure}[!h]
    \centering

\includegraphics[width=8cm,height=6.5cm]{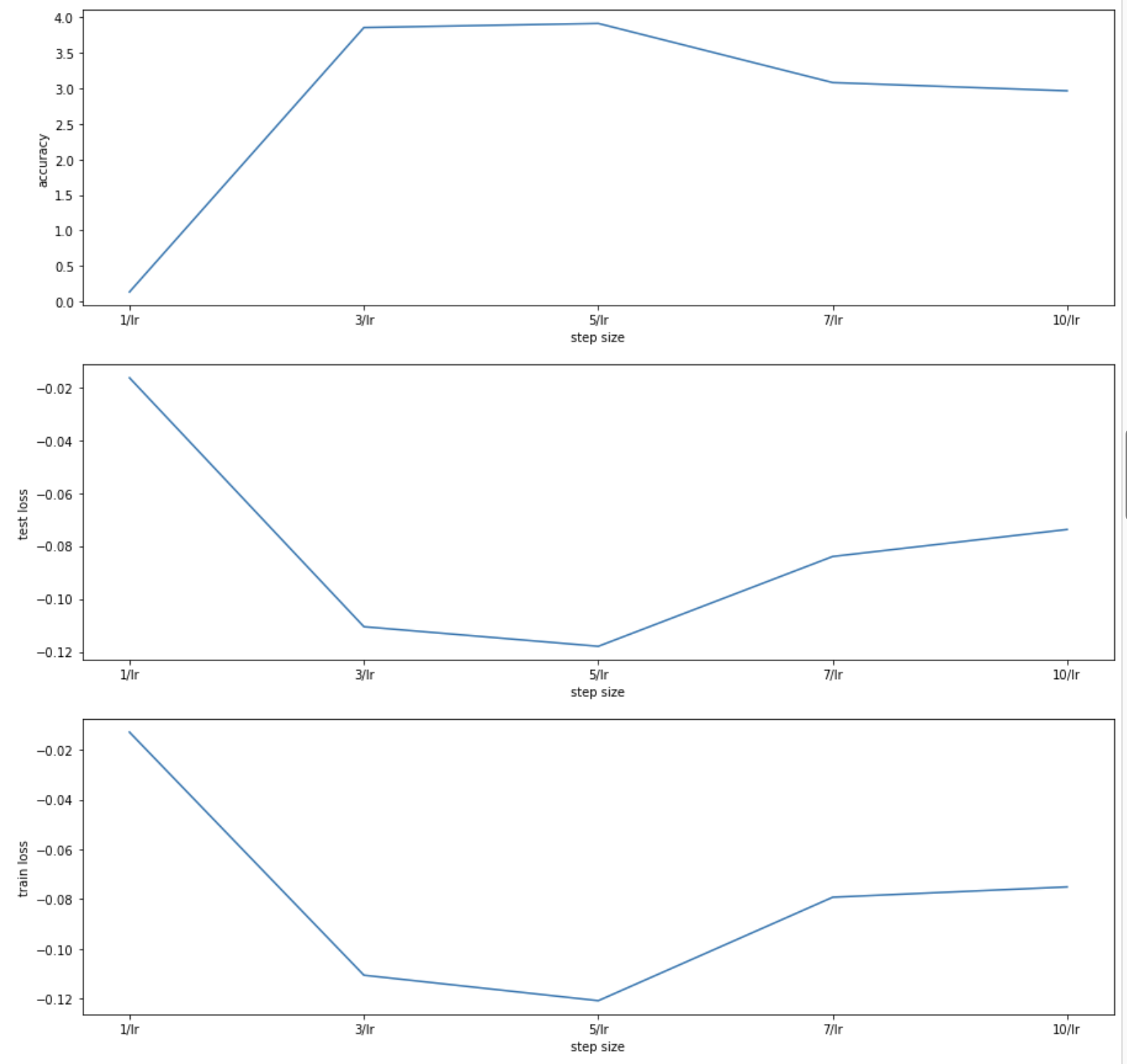}
    
    \caption{Average difference in metrics for different step sizes}
    \label{fig:stepsize}
\end{figure}

We also executed experiments to investigate the effects of step sizes on the performance of our algorithm. Again, we repeated the experiment for various step sizes that are multiples of the reciprocal of the learning rate. Table \ref{table:stepsize} shows the difference between metrics for our algorithm and asynchronous averaged over the training interval. Ideally, for smaller step sizes, our algorithm performs similarly to the synchronous algorithm. As we increase the step sizes, the behavior shifts towards the asynchronous algorithm. The plot \ref{fig:stepsize} shows the relation between step size and the performance of the algorithm.

\begin{table}[h!]
\centering
\begin{tabular}{ |c|c|c|c|c|c| } 
 \hline
 $\frac{(Step size)}{Metric}$ & $\frac{1}{lr}$ & $\frac{3}{lr}$ & $\frac{5}{lr}$ & $\frac{7}{lr}$ & $\frac{10}{lr}$\\ 
 \hline
 Test Accuracy & 0.136 & 3.857 & 3.915 & 3.083 & 2.967\\
 \hline
 Test loss & -0.016 & -0.110 & -0.118 & -0.084 & -0.074 \\
 \hline
 Train loss & -0.013 & -0.110 & -0.121 & -0.079 & -0.075 \\
 \hline
\end{tabular}
\caption{Difference between the metric for our algorithm and asynchronous algorithm averaged over entire training interval for various step sizes and constant batch size of 32. For better performance, difference in accuracy should be positive and that loss should be negative}
\label{table:stepsize}
\end{table}

\subsection{Effect of communication delay}
\begin{figure}[!h]
    \centering

\includegraphics[width=8cm,height=6.5cm]{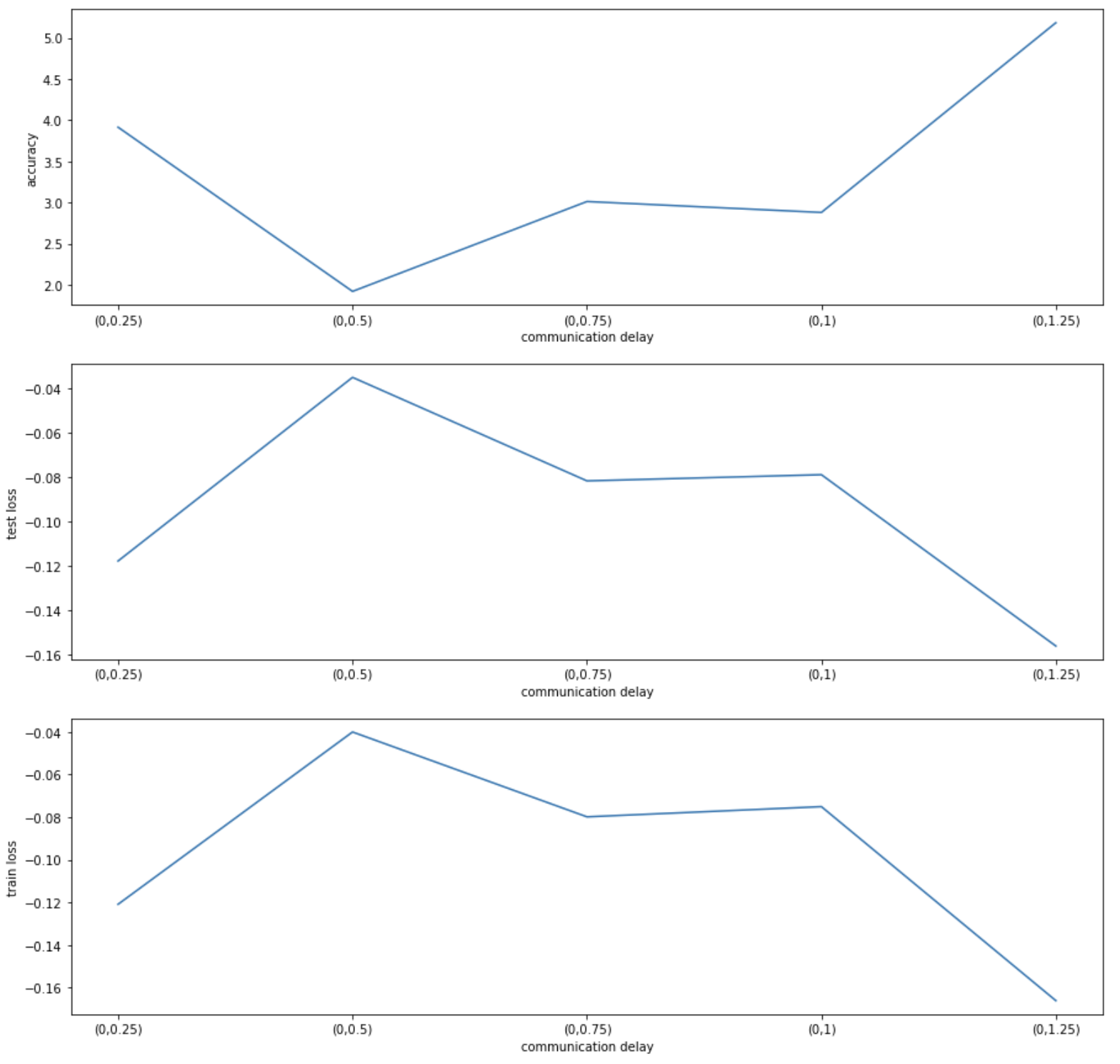}
    
    \caption{Average difference in metrics for different communication delays}
    \label{fig:noise}
\end{figure}

As mentioned in the experiment section, we introduce communication delays by adding random execution delays to the worker. Communication delays are the major reason for the slowness of the synchronous algorithm. However, such a problem is not observed in the asynchronous algorithm, and for our algorithm, we believe that for a good choice of step size and batch size, it should be resilient to communication delay. The table \ref{table:noise} and plot \ref{fig:noise}, show the results from the experiment with varying delay distribution (normal distribution with mean 0 and different standard deviation) and we observe that for all values, our algorithm outperformed the asynchronous version.

\begin{table}[h!]
\centering
\begin{tabular}{ |c|c|c|c|c|c| } 
 \hline
 $\frac{(Mean, Std.)}{Metric}$ & (0,0.25) & (0,0.5) & (0,0.75) & (0,1) & (0,1.25)\\ 
 \hline
 Test Accuracy & 3.915 & 1.920 & 3.012 & 2.879 & 5.184\\
 \hline
 Test loss & -0.117 & -0.035 & -0.081 & -0.079 & -0.156 \\
 \hline
 Train loss & -0.120 & -0.039 & -0.079 & -0.075 & -0.166 \\
 \hline
\end{tabular}
\caption{Difference between the metric for our algorithm and asynchronous algorithm averaged over entire training interval for various delay distribution with constant batch sizes and step size of 32 and 500. For better performance, difference in accuracy should be positive and that loss should be negative}
\label{table:noise}
\end{table}

\section{CONCLUSION}

In summary, we can say that our approach looks to combine the best of both worlds. It uses the asynchronous approach to gain speed and synchronous approach for accuracy. And the switch from asynchronous to synchronous is made using a threshold function which takes the learning rate into account for determining how many parameters get aggregated for synchronization. Based on the extensive experimentation, we can draw the conclusion that for the same time period, our approach leads to better accuracy and lower loss than both synchronous and asynchronous approaches. As batch size decreases, which is the norm for large training datasets, our approach performs even better. The step size is currently defined in terms of  the learning rate and even if the step size is not selected appropriately, our approach will give better results than the synchronous approach.

\section{Future Work}

This approach can be tested on different CPU architectures with smaller memory size and processing power to see what impact it has on the overall performance. Another challenge is to check whether it would work with a non convex loss function since such a function will have multiple local minimas. We can also test this approach with more complex loss functions and model architectures to check its robustness. It can also be tested with larger datasets to see how an increase in dataset size affects performance. Currently, finding the threshold for aggregating parameters is based upon experimental data. However, a good heuristic can be devised which can form a base for selecting aggregation the threshold for different types of models and datasets. Different monotonically increasing functions can also be used to see if the all such functions can be straightaway plugged in without much change in performance. Formal proofs can also be derived for the convergence of the parameters even though the approach is not completely synchronous. 

\bibliographystyle{ACM-Reference-Format}
\bibliography{references}

\end{document}